\documentclass[conference]{IEEEtran}
\IEEEoverridecommandlockouts
\usepackage{cite}
\usepackage{amsmath,amssymb,amsfonts}
\usepackage{algorithmic}
\usepackage{graphicx}
\usepackage{textcomp}
\usepackage{xcolor}
\usepackage{booktabs}
\usepackage{tabularx}
\usepackage{array}

\def\BibTeX{{\rm B\kern-.05em{\sc i\kern-.025em b}\kern-.08em
    T\kern-.1667em\lower.7ex\hbox{E}\kern-.125emX}}
\begin{document}

\title{Dynamic Powergrid Optimization with FPGA-based Neuromorphic Higher-order Ising Machines\\

\thanks{S.M. and C.S.T. would like to acknowledge support from the India National Supercomputing Mission (NSM 1.0). S.C. would like to acknowledge support from the WashU McDonnell Scholars Academy.}
}

\author{\IEEEauthorblockN{Saptarshi Maiti\textsuperscript{1}, Shantanu Chakrabartty\textsuperscript{2}, Chetan Singh Thakur\textsuperscript{1}}
\IEEEauthorblockA{\textsuperscript{1}\textit{Department of Electronic Systems Engineering, Indian Institute of Science, Bangalore, India}\\
\textsuperscript{2}\textit{Department of Electrical and Systems Engineering, Washington University in St. Louis, St. Louis, MO, USA}\\
Email: saptarshimai@iisc.ac.in, shantanu@wustl.edu, csthakur@iisc.ac.in
}}


\maketitle

\begin{abstract}
The advent of large, high-density FPGAs such as the AMD Versal enables efficient hardware implementation of neuromorphic Ising machines, unlocking unprecedented performance for complex combinatorial optimizations. A critical application for this computing paradigm is dynamic powergrid optimization, specifically Distribution Network Reconfiguration (DNR), which seeks to mitigate severe resistive losses and reduce electricity costs by optimizing network switch configurations in real time. This paper presents an FPGA-based neuromorphic higher-order Ising machine for high-speed, adaptive DNR. By directly mapping DNR onto higher-order Ising interactions via a polynomial unconstrained binary optimization (PUBO) formulation, the proposed architecture requires only 179 spins for the IEEE 33-bus network, which represents a 6$\times$ reduction compared to standard quadratic formulations. Implemented on an AMD Versal Premium VP1802 FPGA, the architecture executes parallel spin-group updates every 20 ns and solves 33-, 51-, and 69-bus benchmark networks with median latencies of 1.6-29.3 ms, yielding a 7-58$\times$ speedup over Gurobi to the first optimal solution. Using the solver with expanded network-switching flexibility reduces power loss by 80.71 kW (57.83\%), while adaptive reconfiguration every 10 s under time-varying demand saves 131.16 kWh/day over static daily reconfiguration, highlighting the potential of higher-order Ising hardware for fast, adaptive smart grid management.
\end{abstract}


\section{Introduction}
\label{sec:introduction}

\begin{figure}[t]
\centering
\includegraphics[width=\columnwidth]{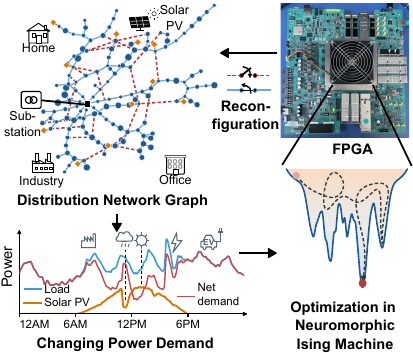}
\caption{Overview of FPGA-based neuromorphic Ising optimization for distribution network reconfiguration under dynamically changing load conditions.}
\label{fig}
\end{figure}

Electric power distribution networks deliver electricity from substations to homes, businesses, and industries through medium- and low-voltage feeders. A significant amount of electricity is dissipated as resistive ($I^2R$) losses in these networks, increasing operating costs and wasting generation capacity. Worldwide, transmission and distribution losses account for approximately 5-20\% of generated electricity~\cite{worldbank2025tdl,eia2023losses,ceer2025losses}. With global electricity generation exceeding 30,000~TWh annually~\cite{iea2025ger}, even a modest 1\% reduction in these losses can save over \$1 billion per year. Delhi, for example, has cut its losses roughly tenfold through grid modernization that also enabled remote network control~\cite{thomas2026delhi}. Unlike building a new infrastructure, optimizing the flow of electricity through such control offers a cost-effective way to further reduce losses. Reconfigured distribution networks in the TEPCO power grid have recently achieved annual loss reductions of up to 6\%~\cite{hagiwara2026screening}.

Distribution networks are generally operated in a radial topology, where each consumer is connected to the substation through a single path. However, alternative power-flow paths can be established by operating sectionalizing and tie switches. Selecting the configuration that minimizes resistive losses while preserving radiality and operating limits is known as distribution network reconfiguration (DNR)~\cite{wu1989network,behbahani2024}. Growing deployment of rooftop solar panels, electric vehicles, and variable loads makes this optimization increasingly important, as the loss-minimizing configuration changes with demand and power generation. Dynamic DNR repeatedly updates the network configuration to adapt to these variations, as illustrated in Fig.~\ref{fig}. Although physical switching is constrained by equipment wear, faster optimization allows grid operators to reassess configurations more frequently and identify opportunities.

Finding the optimal configuration is computationally challenging because the number of possible switch combinations grows combinatorially, while each configuration must satisfy network radiality and electrical constraints. Conventional approaches include branch-exchange heuristics~\cite{civanlar1988,wu1989network}, metaheuristic algorithms~\cite{chiang1990,nara1992}, and mixed-integer optimization~\cite{jabr2012,lavorato2012}. Heuristics can converge to local optima, while stochastic searches often require numerous objective evaluations. Mathematical programming may become computationally expensive for larger networks and extended switching flexibility. These limitations motivate specialized hardware for frequent, high-speed combinatorial optimization. Rooftop PV, electric-vehicle
charging, and flexible loads make the net load vary even faster (clouds can halve the output of a PV system within seconds~\cite{cormode2013}) and can reverse power flows. The optimal configuration therefore changes frequently during the day.

Ising machines offer an alternative computing paradigm by encoding combinatorial problems as an energy function of interacting binary spins~\cite{mohseni2022,lucas2014}. Their stochastic dynamics and local-field updates make them attractive for hardware implementations~\cite{aadit2022}. Recently, DNR has been formulated as quadratic unconstrained binary optimization (QUBO), enabling quantum annealing and Ising-based optimization~\cite{silva2023qubo,Silva2023}. However, the underlying power-flow and radiality expressions naturally contain higher-order interactions, requiring auxiliary variables when reduced to quadratic form. This increases the number of spins and associated hardware resources~\cite{silva2023qubo}. Higher-order Ising machines directly implement multi-spin interactions, avoiding this overhead~\cite{bybee2023,ahsan2026}.

In this work, we develop a compact higher-order DNR formulation implemented on an FPGA accelerator based on the neuromorphic higher-order Ising machine (Fig.~\ref{fig}). The proposed architecture directly implements polynomial interactions with fixed-point arithmetic, interactions, and parallel spin updates. We evaluate the hardware across distribution network benchmarks and investigate its application to expanded switching flexibility and dynamic reconfiguration under variability.

\section{DNR Formulation and Ising Mapping}
\label{sec:formulation}

\subsection{Minimum-Loss Network Reconfiguration}

A distribution network is a graph $G=(V,E)$ of buses $V$ and lines $E$ (Fig.~\ref{fig}). Closed lines form a spanning tree $T\in\mathcal{T}(G)$, the set of spanning trees of $G$, in radial operation. With constant-current approximation~\cite{silva2023qubo}, bus $n$ draws $I_n=(P_n-jQ_n)/V_0$, with $P_n$ and $Q_n$ as net active and reactive demands and $V_0$, the nominal voltage. A closed line $\ell$ with resistance $R_\ell$ carries $I_\ell(T)=\sum_{n\in D_\ell(T)}I_n$, where $D_\ell(T)$ is the set of buses supplied through $\ell$. The minimum-loss configuration is
\begin{equation}
\label{eq:loss}
T^*=\underset{T\in\mathcal{T}(G)}{\arg\min}\;P_{\mathrm{loss}}(T),\quad
P_{\mathrm{loss}}(T)=\sum_{\ell\in T}R_\ell\,|I_\ell(T)|^2.
\end{equation}

\subsection{Compact PUBO Formulation}

Since an Ising machine minimizes an unconstrained energy, minimizing $P_{\mathrm{loss}}$ alone would return a non-radial network. We therefore first rewrite~\eqref{eq:loss} as a PUBO in which binary variables express both the loss and, as penalties, the radiality constraints. We encode $T$ in a binary vector $\mathbf{x}$ using the variables of~\cite{silva2023qubo}. Each chain of degree-two buses is contracted into a lifted path between two reduced vertices; a radial configuration either closes a lifted path or opens exactly one of its lines. Arc variables $e_{uv}$ select the closed paths and orient them away from the root, and cycle-direction variables $d$ enter the constraints that exclude directed cycles. For an intermediate
bus $m$, $p_m=1$ indicates that it is fed from the lower-indexed end of its path. Finally, $z_{uv,n}=1$ indicates that bus $n$ is downstream of arc
$u\!\to\!v$, i.e., its current flows through that path, with $z_{uv,v}\equiv e_{uv}$. We call $\mathbf{x}$ valid if it satisfies the constraints in~\cite{silva2023qubo}; this ensures that it encodes a valid tree $T$.

To keep the loss quadratic, a QUBO formulation uses a separate variable $z_{uv,n}$ for each bus~\cite {silva2023qubo}. We retain $z_{uv,n}$ only for
reduced vertices. Consider an intermediate bus $m$ on another lifted path $(a,b)$ with $a<b$. Bus $m$ is downstream of $u\!\to\!v$ exactly when the
end that feeds it is. This end is $a$ if $p_m=1$ and $b$ otherwise. Hence, without a new variable,
\begin{equation}
\label{eq:compressed_route}
z_{uv,m}=p_m\,z_{uv,a}+(1-p_m)\,z_{uv,b}.
\end{equation}

Let $\delta_{\ell n}(\mathbf{x})=1$ if line $\ell$ carries the current of bus $n$. If $n$ is an intermediate bus on the same lifted path as $\ell$, $\delta_{\ell n}$ is $p_n$ or $1-p_n$, depending on which side of $\ell$ the bus lies. Otherwise, $\delta_{\ell n}$ is given by the $z$
variables of the arcs of that path. So the current through line $\ell$ is  $I_\ell(\mathbf{x})=\sum_n\delta_{\ell n}(\mathbf{x})I_n$. Now from~\eqref{eq:loss},
\begin{equation}
\label{eq:binary_loss}
P_{\mathrm{loss}}(\mathbf{x})=\sum_{\ell\in E}R_\ell
\Bigl|\sum_{n\in V}\delta_{\ell n}(\mathbf{x})\,I_n\Bigr|^2 .
\end{equation}
For a valid $\mathbf{x}$, $P_{\mathrm{loss}}(\mathbf{x})=P_{\mathrm{loss}}(T)$; \eqref{eq:binary_loss} has degree $\leq$ 4. We then expand $\delta_{\ell n}$'s and add constraints as penalties $C_r(\mathbf{x})$~\cite{silva2023qubo}. Each $C_r$ is a polynomial in terms of the binary variables and is zero when its constraint holds. The final PUBO objective is
\begin{equation}
\label{eq:pubo}
H_{\mathrm{PUBO}}(\mathbf{x})=\alpha\,P_{\mathrm{loss}}(\mathbf{x})
+\lambda\sum_r C_r(\mathbf{x}),
\end{equation}
with $\alpha=0.002\,\mathrm{kW}^{-1}$ and $\lambda=2$. For a valid $\mathbf{x}$, $H_{\mathrm{PUBO}}(\mathbf{x})=\alpha P_{\mathrm{loss}}(T)$.
Any violated constraint adds at least $\lambda$, which is larger than the scaled optimal loss. The minimizer of~\eqref{eq:pubo} therefore encodes $T^*$. Efficient PUBO encoding on the benchmark IEEE 33-bus network \cite{wu1989network} reduced the number of spins from 1,074 in QUBO \cite{silva2023qubo} to 179, as summarized in Table~\ref{tab:qubo_pubo_comparison}.

\begin{table}[t]
\centering
\caption{Binary-variable comparison for the IEEE 33-bus network}
\label{tab:qubo_pubo_comparison}
\footnotesize
\setlength{\tabcolsep}{3pt}
\renewcommand{\arraystretch}{1.12}
\begin{tabular}{l|rrrrrr|r}
\hline
\textbf{Method} & $\mathbf{e}$ & $\mathbf{d}$ & $\mathbf{p}$ & $\mathbf{z}$ & $\mathbf{y}$ & \textbf{Aux.} & \textbf{Total} \\
\hline
QUBO~\cite{silva2023qubo} & 24 & 4 & 23 & 577 & 434 & 12 & \textbf{1074} \\
PUBO (ours) & 24 & 4 & 23 & 128 & 0 & 0 & \textbf{179} \\
\hline
\end{tabular}
\vspace{2pt}
\parbox{\columnwidth}{\scriptsize $e$: directed arcs; $d$: cycle directions; $p$: feeding directions; $z$: load-arc routing; $y$: intermediate variables; Aux.: additional auxiliary variables.}
\end{table}

\subsection{Higher-Order Ising Machine}

Substituting $x_i=(1+s_i)/2$, with spins $s_i\in\{-1,+1\}$,
into~\eqref{eq:pubo} and expanding
$H_{\mathrm{Ising}}(\mathbf{s})=H_{\mathrm{PUBO}}\bigl((\mathbf{1}+\mathbf{s})/2\bigr)$
gives
\begin{equation}
\label{eq:spin_hamiltonian}
H_{\mathrm{Ising}}(\mathbf{s})=H_0-\sum_{k=1}^{M}J_k\prod_{i\in S_k}s_i,
\end{equation}
where $H_0$ is a constant and $J_k$ couples the spins in $S_k$. The mapping is exact, so the ground state encodes $T^*$. For the IEEE 33-bus feeder, \eqref{eq:spin_hamiltonian} has 179 spins and 20{,}300 terms of up to order 4. The couplings depend on the line resistances and bus demands via~\eqref {eq:binary_loss}, and on the topology via the penalties. The neuromorphic higher-order Ising machine~\cite{ahsan2026} searches for this ground state stochastically: at each iteration, spin $s_i$ flips if its energy change $\Delta H_i=2s_ih_i$,
with local field $h_i=\sum_{k:\,i\in S_k}J_k\prod_{j\in S_k\setminus\{i\}}s_j$, is below a random noise threshold. Annealing the noise over time drives the spins to near-optimal configurations, which Section~\ref{sec:hardware} implements on FPGA.

\section{FPGA Architecture}
\label{sec:hardware}

\begin{figure}[t]
\centering
\includegraphics[width=\columnwidth]{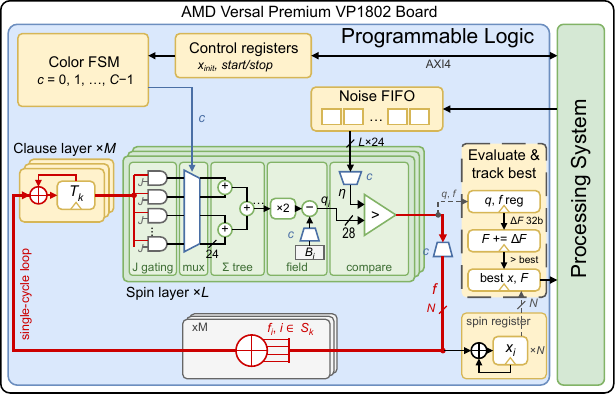}
\caption{Architecture of the FPGA-based neuromorphic higher-order Ising machine, showing the two-layer architecture, interaction mapping, parallel spin-update datapaths, clause-state updates, and evaluation and tracking.}
\label{fig:nhoim_arch}
\end{figure}

\subsection{Architecture Overview}

Evaluating $\Delta H_i$ directly would require recomputing the spin product of every term that contains $s_i$. Instead, we treat each term
of~\eqref{eq:spin_hamiltonian} as a clause and store its sign $c_k=\prod_{j\in S_k}s_j$ in a dedicated clause neuron, which flips whenever
one of its spins flips. So the energy change reduces to a signed sum, $\Delta H_i=2\sum_{k:\,i\in S_k}J_kc_k$. Fig.~\ref{fig:nhoim_arch} illustrates the two-layer architecture: spin neurons compute $\Delta H_i$ and make the stochastic updates, while clause neurons hold the clause states. The spin-clause connections are set at compile time in FPGA logic and registers, so no interconnection data is fetched from memory. The neurons operate in parallel. Spins are partitioned into color groups-no two spins share a clause; their energy changes are therefore independent, and each group is updated simultaneously, as determined by the color controller \cite{ahsan2026}. The architecture comprises $N$ spin registers, $M$ clause registers, $C$ colors and $L$ parallel spin-update lanes.

\subsection{Spin-Update Datapath}

Each clause neuron stores the sign of its term as one bit: $T_k=1$ if $c_k=+1$ and $T_k=0$ otherwise. With fixed-point couplings $J_{i,k}$, spin neuron $i$ computes
\begin{equation}
\label{eq:hw_field}
q_i=2\!\!\sum_{\substack{k:\,i\in S_k\\ T_k=1}}\!\!J_{i,k}-B_i,\qquad
B_i=\sum_{k:\,i\in S_k}J_{i,k}.
\end{equation}
$B_i$ is a precomputed constant. Since $c_k=2T_k-1$, $q_i=\sum_{k:\,i\in S_k}J_{i,k}c_k$, i.e., the value of $\Delta H_i/2$. Each clause gates its coefficient into a parallel adder tree, avoiding multiplication.

Since only one color is updated at a time, at most $L$ flip decisions are needed in parallel, where $L$ is the size of the largest color. We therefore instantiate $L$ spin-update lanes instead of $N$ and time-share them across colors, thereby avoiding a datapath per spin without reducing throughput across the evaluated networks; $L$ ranges from 10 to 22~ (Table \ref{tab:hardware_benchmark}). Each lane flips its spin when $q_i<\eta_i$, where $\eta_i$ is the noise threshold of Section~\ref{sec:formulation} scaled in the same way as $q_i$. The thresholds are generated from the annealing schedule and enter through a rolling FIFO that needs one new threshold per color update. All spins of a color update together, as follows.
\begin{equation}
\label{eq:hw_term_update}
x_i^{+}=x_i\oplus f_i,\qquad
T_k^{+}=T_k\oplus\bigoplus_{i\in S_k}f_i,
\end{equation}
where $x_i=(1+s_i)/2$ is the spin bit. Each clause toggles on at most one flip signal per update, regardless of its order.

The objective tracker incrementally updates the objective energy each update and records the best state. It runs in a separate pipeline stage, so the core sustains one color update per clock without waiting for the objective.

\subsection{FPGA Implementation}

The architecture is implemented on an AMD Versal Premium VP1802 FPGA with 24-bit signed coefficients and 28-bit local fields. Clauses whose quantized coupling is zero are removed. This pruning leaves the quantized Hamiltonian unchanged, and the reference optima still remain reachable in all evaluated networks. As shown in Table~\ref{tab:hardware_benchmark}, pruning reduces the IEEE-69 interaction count from 54,945 to 25,237 (54.1\% fewer), which substantially reduces the combinational logic, without losing the ground state.

\begin{table}[t]
\centering
\caption{FPGA implementation on AMD Versal VP1802 at 50~MHz}
\label{tab:hardware_benchmark}
\footnotesize
\setlength{\tabcolsep}{2pt}
\renewcommand{\arraystretch}{1.15}
\begin{tabular*}{\columnwidth}{@{\extracolsep{\fill}}lccccc@{}}
\hline
\textbf{Network}
& \textbf{Spins}
& \shortstack{\textbf{Terms (pruned)}}
& \shortstack{\textbf{C / L}}
& \textbf{LUTs (\%)}
& \textbf{FFs (\%)} \\
\hline
IEEE-33
& 179 & 20,300 (20,300) & 31/17 & 217.9k (6.48) & 69.5k (1.03) \\
Cairo-51
& 141 & 49,187 (44,699) & 48/10 & 594.3k (17.68) & 174.6k (2.60) \\
IEEE-69
& 385 & 54,945 (25,237) & 51/22 & 221.7k (6.60) & 32.2k (0.48) \\
\hline
\end{tabular*}
\end{table}

All three network-specific implementations were run at 50~MHz, sustaining one parallel color update every 20~ns. Despite having fewer spins, Cairo-51 requires more LUTs due to its denser spin-interaction connectivity. The complete implementations occupy only 6.48-17.68\% of available LUTs, with no DSP or URAM usage. AXI4-based interface provides control and result readout. The results demonstrate the resource efficiency of fixed-point higher-order Ising computation across different network sizes (Table~\ref{tab:hardware_benchmark}).

\section{Experimental Results}
\label{sec:results}

\begin{figure}[t]
\centering
\includegraphics[width=\columnwidth]{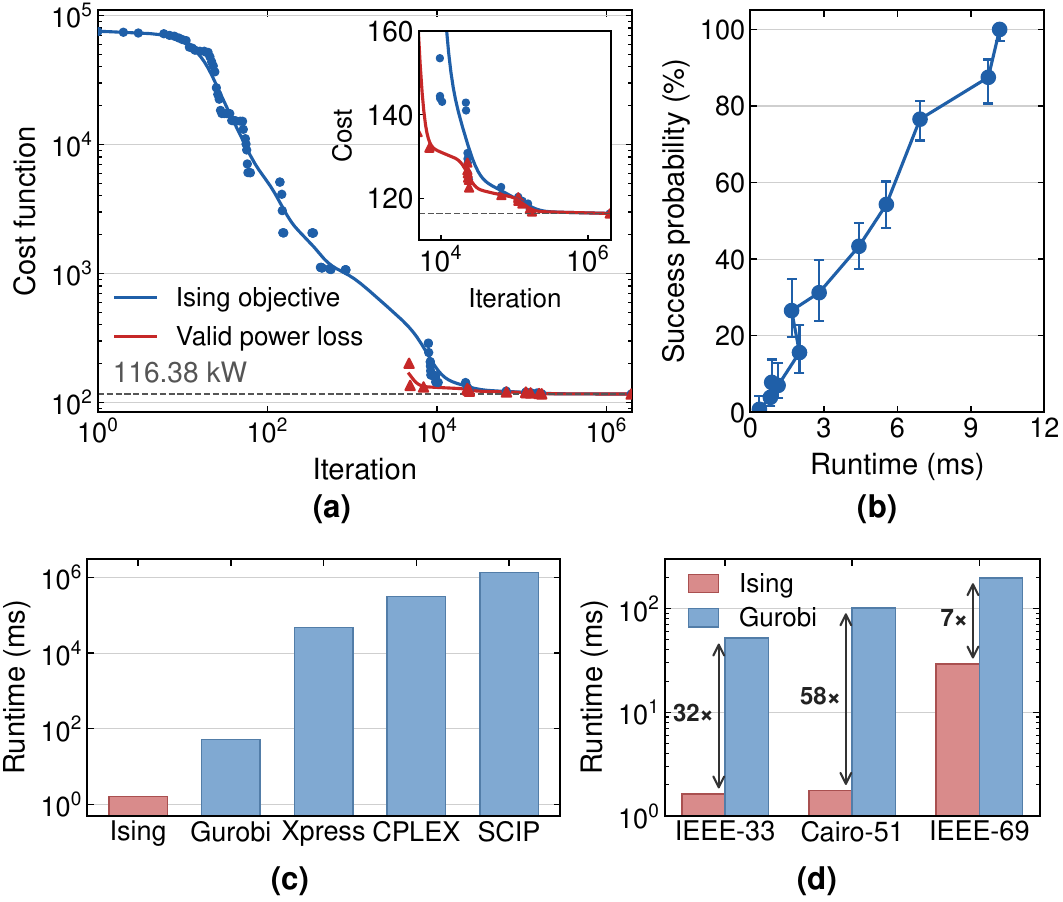}
\caption{Ising optimization and benchmarking results. (a) Evolution of the Ising objective (scaled) and valid power loss for IEEE-33, with an inset showing convergence near the optimum. (b) Success probability versus median first-hit latency for different annealing budgets; error bars indicate Wilson 95\% confidence intervals. (c) IEEE-33 runtime comparison with classical solvers \cite{Silva2023} (d) Median latency of Ising and Gurobi across three distribution networks.}
\label{fig:nhoim_results}
\end{figure}

We evaluate the Ising solver on IEEE 33-bus~\cite{wu1989network}, Cairo 51-bus~\cite{Atef2025}, and IEEE 69-bus~\cite{Baran1989} networks on the VP1802 FPGA. Each independent run starts from a random spin configuration, and success is defined by reaching the independently established reference optimum of the constant-current loss formulation. Latencies correspond to FPGA Ising-core latencies, excluding communication and readout overheads.

\subsection{Optimization Performance}

Fig.~\ref{fig:nhoim_results}(a) illustrates the optimization trajectory for IEEE-33. As the unconstrained Ising objective improves, feasible radial configurations emerge, and the best feasible loss converges to the reference optimum of 116.38~kW. The annealing-budget sweep in Fig.~\ref{fig:nhoim_results}(b) reveals a reliability-latency trade-off: slower annealing makes it more likely to hit the ground state at the cost of longer runtime. The annealing schedules are chosen to give the best 99\% time-to-solution \cite{ahsan2026}, accounting for both runtime and success probability under independent restarts.
Fig.~\ref{fig:nhoim_results}(c)-(d) compares the Ising machine with conventional solvers~\cite{Silva2023}, where the Ising machine was the fastest. Across IEEE-33, Cairo-51, and IEEE-69, Ising achieves median successful first-hit latencies of 1.631, 1.765, and 29.302~ms, respectively, compared with 51.90, 101.67, and 198.74~ms for Gurobi, corresponding to approximately 32$\times$, 58$\times$, and 7$\times$ shorter first-hit times. Gurobi 13.0.3 was executed on an AMD EPYC 9254 processor using the native constrained DNR formulation. Table~\ref{tab:dnr_timing_comparison} further compares our FPGA-based neuromorphic higher-order Ising machine with other reported classical, quantum, and learning-based approaches for DNR.

\subsection{Switching Flexibility and Dynamic Reconfiguration}

Beyond solver speed, we evaluate our higher-order Ising machine-based DNR solver at the application level, where fast solutions can allow distribution grids to use more switches for flexibility and respond to changing conditions. The Ising machine minimizes the constant-current objective~\eqref{eq:loss}, and the selected configurations are evaluated with AC power flow \cite{Shirmohammadi1988}.

\begin{figure}[t]
\centering
\includegraphics[width=\columnwidth]{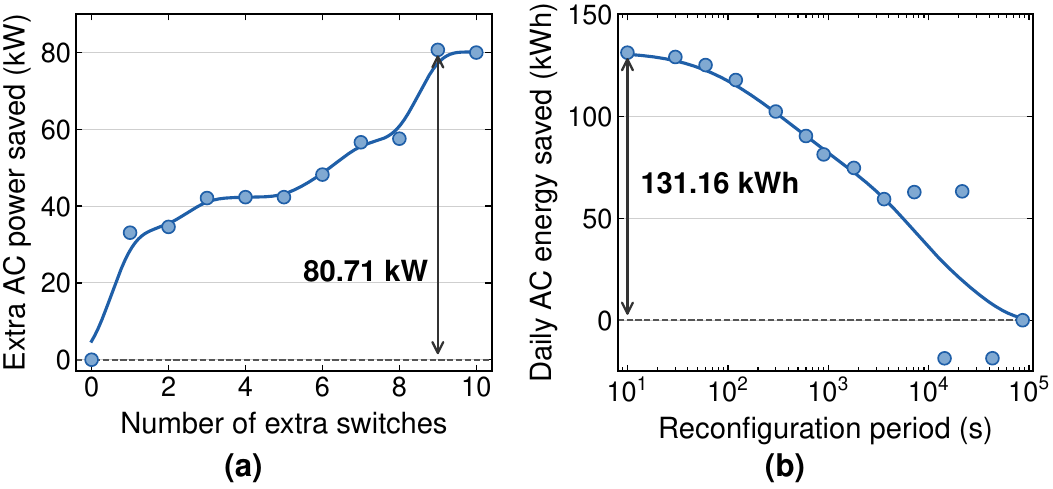}
\caption{Application-level DNR optimization using an Ising machine. (a) AC power-loss savings in the IEEE 33-bus network as additional candidate switches increase network flexibility. (b) AC energy savings versus reconfiguration interval under time-varying power demand throughout the day.}
\label{fig:dnr_application_results}
\end{figure}

Fig.~\ref{fig:dnr_application_results}(a) shows the benefit of adding candidate switches to the IEEE-33 network~\cite{Ehsanbakhsh2023}. Without extra switches, the selected configuration has an AC loss of 139.55~kW. Ten additional switches expand the set of feasible radial configurations exponentially, with the Ising machine regularly finding optimal configurations, reducing the AC power loss to 58.84~kW, an additional saving of 80.71~kW (57.83\%). We further evaluate Ising-based dynamic reconfiguration simulated over a 24-hour profile with changing power demands~\cite{sengupta2010oahu}, and network configurations repeatedly optimized at different periods using the solver as illustrated in Fig.~\ref{fig:dnr_application_results}(b). Relative to once-daily reconfiguration, hourly optimization saves 59.37~kWh/day, increasing to 131.16~kWh/day with 10-s reconfiguration.

\begin{table}[tbp]
\centering
\caption{Comparison of DNR Solvers on IEEE 33-Bus Network.}
\label{tab:dnr_timing_comparison}
\scriptsize
\setlength{\tabcolsep}{1.2pt}
\renewcommand{\arraystretch}{1.08}

\begin{tabularx}{\columnwidth}{
@{}
>{\raggedright\arraybackslash}X
c
>{\raggedright\arraybackslash}p{0.20\columnwidth}
r
@{}
}
\hline
\textbf{Method} &
\textbf{Study} &
\textbf{Implementation} &
\textbf{Latency} \\
\hline

Graph-based Genetic Algorithm
& \cite{Raut2017}
& MATLAB
& 6\,s
\\

Chaotic Stochastic Fractal Search
& \cite{Tran2020}
& MATLAB / CPU
& 49.59\,s
\\

Particle Swarm Optimization
& \cite{kahouli2021pso}
& MATLAB
& 120.25\,s
\\

QUBO / Hybrid Quantum Annealing
& \cite{Silva2023}
& D-Wave hybrid
& 60\,s
\\

Mixed-Integer Quadratic Programming
& \cite{Silva2023}
& Xpress / CPU
& 48\,s
\\

Mixed-Integer Quadratic Programming
& \cite{Silva2023}
& CPLEX / CPU
& 316\,s
\\

Improved Simulated Annealing
& \cite{simeon2024isahc}
& OpenDSS / CPU
& 4.25\,s
\\

Mixed-Integer Linear Programming
& \cite{Graine2026}
& Gurobi / CPU
& 425\,ms
\\

Quantum Approximate Optimization Algorithm
& \cite{Bosmediano2026}
& Qiskit Aer
& 1.36\,s
\\

Deep Reinforcement Learning
& \cite{Ghaemipour2026}
& CPU / GPU
& 59.1\,ms
\\

Hybrid Simulated Annealing-Tabu Search
& \cite{bautista2026hybrid}
& Python
& 868\,ms
\\

\hline
\textbf{Neuromorphic Higher-order Ising Machine}
& \textbf{This work}
& \textbf{FPGA}
& \textbf{1.63\,ms}
\\
\hline

\end{tabularx}

\vspace{2pt}

\end{table}

\section{Conclusions}

This work demonstrates the potential of dedicated higher-order Ising hardware for fast, resource-efficient optimization of electrical distribution networks. By eliminating auxiliary variables and exploiting distributed parallel computation, the proposed architecture renders large-scale combinatorial searches feasible in milliseconds on a single FPGA. Such fast, low-overhead optimization enables grid operators to explore larger configuration spaces and reassess network conditions more frequently, improving infrastructure utilization without requiring extra generation capacity. Beyond power loss minimization, this approach can be extended to other grid operations like fault isolation, service restoration, contingency reconfiguration, and microgrid islanding, where the Ising accelerator can enable faster decision-making and shorter response times.


\bibliographystyle{IEEEtran}
\bibliography{bibliography}

\end{document}